%% file: neurips_2026.tex
\documentclass{article}

\PassOptionsToPackage{numbers, compress}{natbib}

\usepackage[position,final]{neurips_2026}

\usepackage[utf8]{inputenc} 
\usepackage[T1]{fontenc}    
\usepackage{hyperref}       
\usepackage{url}            
\usepackage{booktabs}       
\usepackage{amsfonts}       
\usepackage{nicefrac}       
\usepackage{microtype}      
\usepackage{xcolor}         
\usepackage{enumitem}
\usepackage{amsmath}
\usepackage{amsthm}

\usepackage{algorithm}       

\usepackage{algpseudocode}  
\usepackage{amsmath}         
\usepackage{amssymb}         
\usepackage{wrapfig}
\usepackage{amsfonts}        
\usepackage{graphicx}
\usepackage{listings}
\usepackage{xcolor}
\usepackage{multirow}
\usepackage[most,skins,theorems]{tcolorbox}
\tcbset{
  aibox/.style={
    width=\linewidth,
    top=8pt,
    bottom=4pt,
    colback=blue!6!white,
    colframe=black,
    colbacktitle=black,
    enhanced,
    center,
    attach boxed title to top left={yshift=-0.1in,xshift=0.15in},
    boxed title style={boxrule=0pt,colframe=white,},
  }
}
\newtcolorbox{AIbox}[2][]{aibox,title=#2,#1}

\newcommand{\eg}{\textit{e.g.}}

\newcommand{\ie}{\textit{i.e.}}

\title{Superplatforms Are Strategically Compelled to Counteract AI Agents}

\author{%
  Jianghao Lin \\
  Shanghai Jiao Tong \\
  University \\
  \texttt{linjianghao@sjtu.edu.cn}
  \And
  Jiachen Zhu$^*$ \\
  Shanghai Jiao Tong \\
  University \\
  \texttt{gebro13@sjtu.edu.cn}
  \And
  Zheli Zhou \\
  Shanghai Jiao Tong \\
  University \\
  \texttt{zlz114514@sjtu.edu.cn}
  \AND
  Yunjia Xi \\
  Shanghai Jiao Tong 
  University \\
  \texttt{xiyunjia@sjtu.edu.cn}
  \And
  Weiwen Liu \\
  Shanghai Jiao Tong 
  University \\
  \texttt{wwliu@sjtu.edu.cn}
  \AND
  Yong Yu \\
  Shanghai Jiao Tong 
  University \\
  \texttt{yyu@sjtu.edu.cn}
  \And
  Weinan Zhang\thanks{Corresponding author.} \\
  Shanghai Jiao Tong 
  University \\
  \texttt{wnzhang@sjtu.edu.cn}
}
\begin{document}

\maketitle

\begin{abstract}
This position paper argues that superplatforms are strategically compelled to counteract, constrain, or even disrupt AI agents as these agents emerge as competing digital gatekeepers. Superplatforms have built their business models on controlling user attention through targeted advertising, algorithmic content curation, and centralized access to digital services. However, LLM-driven AI agents threaten this model by acting on behalf of users, bypassing platform-controlled interfaces, reducing exposure to advertisements, and potentially becoming the new entrance for digital traffic. Drawing on gatekeeping theory, we analyze why this shift creates a structural conflict between superplatforms and AI agents: whoever controls the user’s primary interface controls information flow, user data, and monetization opportunities. We then examine why common responses, such as proprietary agents and API gating, are insufficient against general-purpose and GUI-based agents, motivating the possibility of more proactive countermeasures. Finally, we outline a taxonomy of such platform-initiated countermeasures and identify the technical challenges they raise. We do not advocate for adversarial disruption; rather, our goal is to surface this emerging tension early and encourage research toward open, collaborative, and user-centric agent ecosystems.
\end{abstract}

\input{icml2026/text/introduction}

\input{icml2026/text/why_attack}
\input{icml2026/text/taxonomy}
\input{icml2026/text/impact}
\input{icml2026/text/alternative_view}

\input{icml2026/text/conclusion}

\section*{Acknowledgement}
The Shanghai Jiao Tong University team is
partially supported by National Natural Science Foundation of China (624B2096, 72595872, 72542012, 62322603, 62502310).

\bibliographystyle{ACM-Reference-Format}
\bibliography{example_paper}
\newpage
\appendix
\input{icml2026/text/limitation}
\input{icml2026/text/ethic}
\input{icml2026/text/appendix_experiments}

\end{document}

%% file: icml2026/text/introduction.tex
\section{Introduction}

\begin{figure*}[t]
  \centering
  \includegraphics[width=1.0\textwidth]{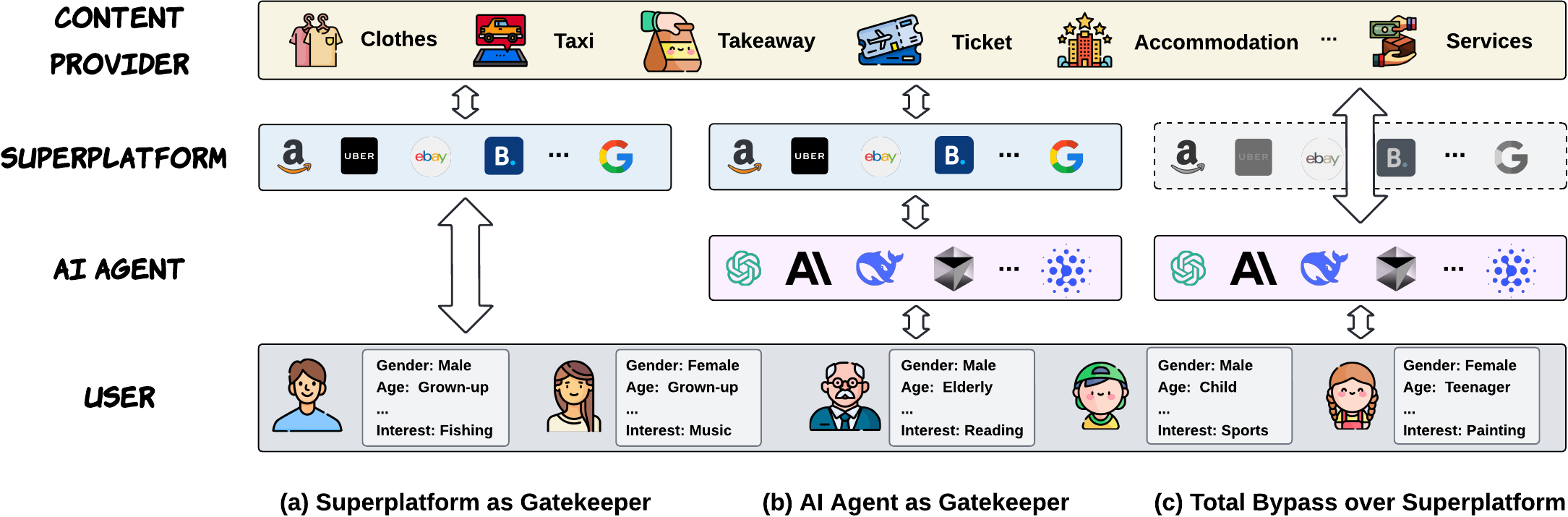}
  \caption{Three paradigms of user interacting with Internet services (\ie, content providers). 
  (a) In the pre-agent age, superplatforms serve as the gatekeeper between the users and content providers. 
  (b) The emerging AI agents intermediate between the users and superplatforms, becoming new gatekeepers. 
  (c) Given the agent-as-gatekeeper fact, content providers can simply bypass superplatforms and offer their Internet services directly to agents. 
  }
  \vspace{-13pt}
  \label{fig: Introduction}
\end{figure*}

The evolution of the Internet has given rise to superplatforms. 
We define a \emph{superplatform} as a platform function that serves as the primary gateway through which users access a broad ecosystem of third-party information or services, while controlling traffic allocation, user interactions, and monetization.  \eg, Google for web search and Amazon for e-commerce. 
As shown in Figure~\ref{fig: Introduction}(a), they leverage their scale to amass and exploit user data across services, intermediating user interactions with centralized controls~\cite{russo2021future}. 
As the digital traffic entrance, superplatforms can build comprehensive user profiles and tightly gatekeep what content or options users can see, all to maximize engagement and monetization~\cite{russo2021future,kapoor2025resist}. 
Substantial revenues are made via \textbf{user-attention-based monetization} by capturing and retaining user attention through curated content, then monetizing this attention by selling prioritized visibility -- such as top-ranking search results or featured product placements -- to the highest bidders.

While user-attention-based monetization requires users to personally engage with the platform through direct interactions, AI agents driven by large language models (LLMs) offer an alternative path to free user attention with autonomy across diverse platforms and tasks~\cite{yang2025survey,dorri2018multi}. 
Such agents can reason on user instructions, make decisions, and execute goals with minimal human oversight towards the operational goals~\cite{zhang2024agentic,zhu2026turingtestscreenbenchmark}. 
Unlike a single-site chatbot, a general-purpose agent can interact with multiple applications or websites, performing actions on behalf of a user (\eg, browsing sites, managing apps, or executing transactions) in an integrated manner. 
This capability promises to free users from manual, platform-by-platform interactions, allowing an agent to carry out the user’s intent directly across the Internet. 

Consequently, fundamental tension underlies the relationship between superplatforms and such AI agents: \textbf{the conflict between user-attention-based monetization versus user-attention-free agent autonomy}. 
Superplatforms monetize user attention through targeted advertising and algorithmic content curation by steering and prolonging user engagement within their walled gardens. 
Nevertheless, AI agents flip this script from two aspects: 
\begin{itemize}[leftmargin=10pt]
    \item AI agents free the user attention and serve the user’s goals first, potentially retrieving information or fulfilling tasks in the most efficient way possible, regardless of any platform’s preferred content or ads. 
    As shown in Figure~\ref{fig: Introduction}(b), an agent navigates directly to the information or service the user requests, skipping over ad-laden feeds and ignoring algorithmic recommendations, \ie, AI agents become the new gatekeepers. 
    \item Moreover, not only do AI agents bypass ads and optimize for user goals, but also, the companies behind these agents (\eg, OpenAI and Anthropic) are building platforms that can redirect user intent, becoming the new entrance for digital traffic. 
    As shown in Figure~\ref{fig: Introduction}(c), AI agents can collect, filter, and rerank collected items from different data sources before ultimately presenting them to users, which is a potential new source of commercial benefits on the Internet.
\end{itemize}
Hence, superplatforms are at risk of being disintermediated by a new breed of algorithmic intermediaries — AI agents as meta-platforms. 
This threatens to erode the “captive audience” model that powers superplatforms' profits. 
This analysis leads us to our core position:

\begin{AIbox}{\textbf{Core Position}}
\textbf{Superplatforms are strategically compelled to counteract, constrain, or even disrupt AI agents}. This imperative arises not only because agents can bypass user-attention-based monetization, but also because they are poised to become the next dominant gatekeepers—intermediating user intent, controlling information flow, and threatening to displace superplatforms as the primary channels for digital traffic.
\end{AIbox}
It is crucial to clarify that \textbf{this position is not an endorsement of such actions}. Rather, it is an analytical forecast of a trend we believe requires urgent academic and industrial attention at the emerging age of AI agents.

The remainder of this paper is constructed as follows. 
In Section~\ref{sec:why attack}, we discuss why superplatforms face an imperative to counteract AI agents. 
Specifically, we are the first to apply the gatekeeping theory to analyze the fundamental threats posed by AI agents to superplatforms, revealing that a rational strategic response is to proactively constrain or disrupt them. 
In Section~\ref{sec:attacks}, we summarize the potential countermeasures for superplatforms to employ against AI agents, forming an unexplored new technical direction tailored for such defensive strategies.
Then, we discuss the potential impact of this paper in Section~\ref{sec:impact} and conclude this paper in Section~\ref{sec:conclusion}.
Limitations and ethical considerations are presented in Section~\ref{sec:limitation} and \ref{sec:statement}, respectively.
Alternative views are discussed in Appendix~\ref{sec:alter view}.


%% file: icml2026/text/why_attack.tex
\section{The Strategic Imperative: A Gatekeeping Analysis of the Conflict}
\label{sec:why attack}

This section casts comprehensive discussions about why superplatforms face a strategic imperative to counteract AI agents. 
We first apply the gatekeeping theory to analyze the fundamental threats that AI agents pose to superplatforms.
Then, we discuss the strategic options available to superplatforms in response to the rise of AI agents, ultimately demonstrating that counteracting these agents emerges as a rational, and perhaps inevitable, strategic choice.

\subsection{Threat Analysis with Gatekeeping Theory}

The business model of superplatforms can be formulated by \textbf{the gatekeeping theory}~\cite{lewin1943forces,barzilai2009gatekeeping,shoemaker2009gatekeeping}. 
They act as gatekeepers of user attention and monetizing that attention via advertising. 
The economic engine of superplatform gatekeepers are underscored by two interlocking factors:
\begin{itemize}[leftmargin=10pt]
    \item \textbf{User Traffic Control.} 
    Gatekeepers hold the authority to determine which content is ultimately presented to users, akin to a gatekeeper deciding who is permitted entry. 
    This control over exposure decisions directly influences which advertisements are seen by users, thereby dictating the allocation of advertising revenue, \ie, user-attention-based monetization. 
    For example, the advertising revenue of Google reaches \$264 billion in 2024~\cite{statista_google_ads}, underscoring how dominant their user-attention-market monopolies have become.
    \item \textbf{User Data Ownership.} 
    Gatekeepers possess exclusive access to users' personalized real-world interaction data, analogous to a gatekeeper being privy to the preferences of a household's residents. This data ownership enables the development of highly effective recommendation and online advertising algorithms tailored to user behaviors and preferences. 
    For example, the averaged number of daily user transactions on Amazon has reached 12.87 million~\cite{landingcube_amazon_stats}, which is in turn used to improve Amazon's online recommendation and advertising.
   
\end{itemize}

Notably, this business model depends on \textbf{users interacting directly with the platform’s interfaces} where ads and promoted content can be inserted (\eg, search results, social feeds, e-commerce products). 
General-purpose AI agents directly threaten this revenue model by altering how users find information and consume content. An effective AI agent acts as an alternative intermediary – one that works for the user rather than the platform. Instead of a person manually scrolling through a feed peppered with ads, a personal AI agent could retrieve the specific content or answers the person wants, bypassing the platform’s ad-laced discovery process. 
For instance, rather than using Google in the usual way (\eg, enter query → view page of links with sponsored results → click sites with ads), a user might ask an AI agent a complex question and receive a synthesized answer drawn from multiple sources, without ever seeing Google’s ads or clicking those top sponsored links. 
\textit{This is empirically supported by the most up-to-date evidence: Google’s global search market share dropped below 90\% for the first time since 2015 in the era of AI agents}~\cite{campaignasia_llm_search}.

As shown in Figure~\ref{fig: Introduction}, \textbf{AI agents are gradually becoming new gatekeepers building on the existing superplatform gatekeepers, and may even bypass them completely.} 
Agents usurp the traffic flow and user entrance from various different superplatforms, threatening from the following aspects:
\begin{itemize}[leftmargin=10pt]
    \item AI agents, acting on behalf of users, often prioritize efficiency and relevance over exposure to advertisements. This behavior disrupts the traditional ad-centric revenue models of superplatforms, leading to reduced ad exposures \& conversions, altered user journeys, and shifted decision-making behaviors.
    
    \item Beyond disrupting advertising, AI agents have the potential to become new gatekeepers of digital information and services. They can not only mediate all digital interactions and determine which contents users engage with, but also own the real-world user feedback data, enabling their providers to refine agentic services and further entrench their positions against superplatforms.

    \item Whoever controls the traffic entrance owns the money. Once users become more reliant on AI agents for daily tasks from shopping to news reading, why should advertisers (\eg, e-commercial merchants, and website owners) pay superplatforms for the ad exposure? 
    They can directly bypass the superplatforms and pay the agent service companies for better product exposure and more guaranteed conversions. 
    The total income of these superplatforms will therefore significantly drop in the long run.
\end{itemize}

In summary, the emerging AI agents jeopardize the gatekeeping power and ad-centric profits of superplatforms. 
\textbf{A secret war has begun between superplatforms and AI agents for the control of gatekeeping powers.}
This sets up strong economic motivations for superplatforms to protect their control, resisting any technology that might divert users away from their curated experiences. 
Their market valuations (usually trillions of dollars collectively) are premised on the continuation of surveillance-advertising models. 
Any disruption to gatekeeping-empowered user-attention-based monetization is a direct threat to their bottom line. 

\subsection{Evaluating Strategic Countermeasures}
\label{sec:countermeasures}

Superplatforms have to take actions in response to the rise of AI agents. 
In this section, we discuss the possible strategic countermeasures for superplatforms from the three aspects, ultimately leading to the conclusion that proactive, and even adversarial, countermeasures may become a rational choice.

\subsubsection{Development of Proprietary AI Agents}

Superplatforms are investing in their own AI agents to retain the gatekeeping control over users' attention and data, \eg, Amazon’s shopping assistant and Uber’s ride-booking agent. 
While these in-house agents are deeply integrated within their own ecosystems, they lack the core capability that defines truly useful AI agents: the ability to perform cross-platform, out-house-conditioned tasks. 
Each agent is isolated—Amazon’s agent can only shop, Uber’s agent can only hail rides, and Booking.com’s agent can only reserve travel. 
But complex user goals, such as booking a ride from the airport to the destination at a certain time after confirming flight and hotel details, require integration across these services. 
No single platform-specific agent can perform such orchestration without access to external platform data. 
In this sense, general agents (\eg, Apple Intelligence) are better choice to displace superplatforms as the primary user interface for cross-platform tasks. 

It is intuitive to conduct multi-agent collaboration across platforms to complete a task.
However, the fundamental question arises — who should be the ultimate gatekeeper of user interactions? 
Once the user chooses an agentic service as the interface, the problem reduces back to the conflict of one general-purpose AI agent against other superplatform services. 
Therefore, developing proprietary AI agents is necessary, but it cannot prevent agents (especially operating system providers like Apple) from displacing and threatening superplatforms in terms of gatekeeping powers.

\subsubsection{API Gating and Pricing}

Another strategy for superplatforms to resist external AI agents is to restrict third-party access to data and functionality unless it occurs strictly on their own terms. 
This often takes the form of API gating and monetization by charging developers or agent providers for access to core services, such as search results, user data, or transaction execution. 
This tactic has been widely adopted: platforms like Google, X, and Reddit have imposed steep API fees, effectively discouraging independent AI usage and making revenues from external invocations. 
However, this strategy is only effective against a specific class of agents - those that rely on API-based function calling to operate, \ie, API agents.

API gating offers little to no protection against GUI agents, which interact with platforms by simulating user behaviors at the interface level, \eg, clicking, scrolling, and reading on-screen elements, just as a human would. 
Since these agents do not rely on formal APIs by imitating user operations, they are far more difficult for platforms to detect, distinguish, or block. 
As a result, while API gating may successfully limit structured access to data for API agents, it does nothing to prevent GUI agents from navigating and controlling applications visually, making it a partial and ultimately insufficient defense mechanism in the broader battle against agent-based disintermediation.

\subsubsection{Strategically Proactive Disruption}

When proprietary development and API gating prove insufficient to preserve the gatekeeping power to users, superplatforms may be left with what appears to be an increasingly rational strategy: proactively employing disruptive tactics against AI agents, particularly those relying on GUI-level operations, since API gating strategy is effective enough for API agents.

GUI agents pose the greatest threat to superplatform dominance because they bypass traditional access points such as APIs, instead using visual understanding and user-simulated operations to navigate web or mobile interfaces autonomously. 
To combat this, superplatforms can turn to adversarial tactics aimed at degrading agent performance or obstructing their ability to function effectively. 
Such adversarial designs exploit the brittleness of visual perception models and undermine the agent's decision-making loop, forcing agents into failure modes without triggering alarms for human users. 

In this light, from a superplatform's perspective, deploying adversarial tactics may be framed not as a malicious act, but as a necessary self-protection strategy. 
As agents grow in capability and autonomy, and as they increasingly intermediate user-platform interactions, employing such tactics proactively could be seen as the last effective strategy for superplatforms to preserve their central role as information and transaction gatekeepers in the digital ecosystem.

%% file: icml2026/text/taxonomy.tex
\section{Potential Countermeasures: Taxonomy and Analysis of Challenges}
\label{sec:attacks}

Given the strategic rationale above, it becomes crucial to systematically understand the technical nature of these potential disruptive tactics. This section develops a taxonomy to analyze the specific forms these countermeasures might take.
Since API agents can be well constrained by API gating and pricing, this section focuses on potential countermeasures targeting GUI agents, which can bypass platforms more easily. \textit{To systematically analyze these potential actions, we borrow the framework of adversarial attacks from the AI security literature.} We will discuss these potential countermeasures through the lens of four standard dimensions: (1) their objectives (Attack Goals), (2) the knowledge required (Attacker Knowledge), (3) their visibility to the user (Attack Visibility), and (4) their timing (Attack Timing). 

For each dimension, we first review the literature on adversarial attacks against agents, and then analyze the unique characteristics of countermeasures that could be initiated by superplatforms. Finally, we summarize the challenges and future directions for this emerging area of research.



\subsection{Attack Goals}

\begin{figure*}[t]
  \centering
  \includegraphics[width=1.0\textwidth]{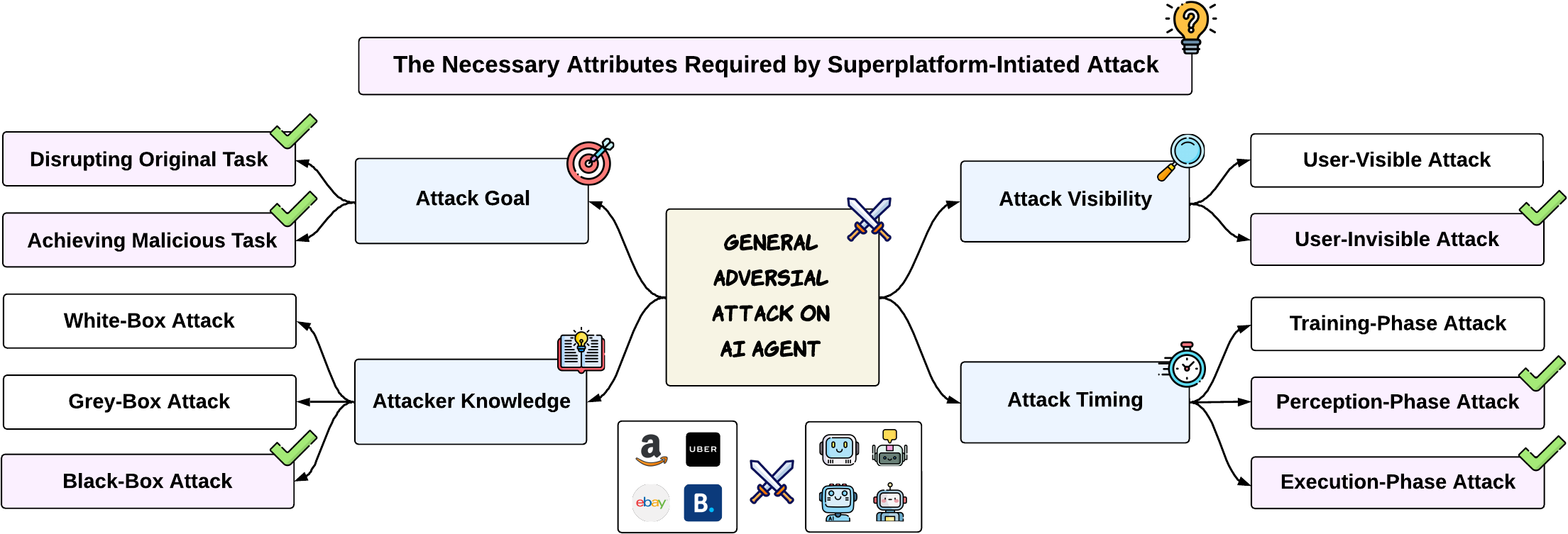}
  \caption{The taxonomy of general adversarial attacks on AI agents, as well as the necessary attributes for such superplatform-initiated countermeasures.
  }
  \vspace{-7pt}
  \label{fig: Attack}
\end{figure*}

\paragraph{Literature Review.}
The objectives of adversarial attacks can vary widely, from achieving specific malicious outcomes to merely disrupting the agent's valid operation. 
The attack goals consist of the following two perspectives:
\begin{itemize}[leftmargin=10pt]
    \item \textbf{Disrupting Agent's Original Task.} 
    These attacks aim to prevent the agent from completing its legitimate, assigned tasks, regardless of additional malicious tasks from attackers~\cite{zhang2024attacking,chen2025aeia}.
    For example, causing inconvenience, operational delays, or denial of service. 
    \citet{zhang2024attacking} demonstrate that adversarial pop-up windows can disrupt agent decision-making, leading to a significant decrease in task success rates. 
    \item \textbf{Achieving Attacker's Malicious Task.} 
    In this scenario, attackers aim to achieve specific malicious results, such as unauthorized actions, data exfiltration, or generating harmful content, no matter if the agent's original task succeeds or not~\cite{chen2025fineprint,xu2024advwebcontrollableblackboxattacks}.
    For instance, AdvWeb~\cite{xu2024advwebcontrollableblackboxattacks} proposes to redirect a stock-trading agent to purchase unintended stocks or perform erroneous bank transactions via adversarial injection. 
\end{itemize}

\paragraph{Application by Superplatforms.}
When superplatforms employ such adversarial tactics, their objectives (\ie, "attack goals" in the context of this taxonomy) are fundamentally aligned with their business interests.
\textit{These goals are not about traditional malicious outcomes like unauthorized access or data exfiltration from the user's perspective, but rather about secretly steering the agent away from its user-defined task or towards superplatform-desired actions.} 
For instance, superplatforms might aim to prevent the agent from successfully completing the user task, \eg, making it repeatedly click or scroll the screen without any meaningful purpose. 
Degenerated performance of AI agents would push users back to superplatforms.
Moreover, superplatforms can redirect the agent to promote specific products for targeted advertising or service improvement. 
This could involve subtle manipulation of certain visible UI elements to draw the agent's attention and prompt a specific action, even if it deviates from the user's explicit instructions. 
These attack goals are distinct from existing research works because they are not designed to break the agent in a destructive way, but rather to manipulate its interpretation of the environment to serve the superplatform's economic or strategic interests.

\subsection{Attacker Knowledge}

\paragraph{Literature Review.}
The effectiveness of adversarial attacks usually correlates with the attacker’s knowledge about the victim agent, including its backend foundation model, user instructions, and operational tasks. 
This dimension ranges from complete access to the victim's internal parameters \& explicit user instructions to limited knowledge with only observable GUI inputs and action outputs.
\begin{itemize}[leftmargin=10pt]
    \item \textbf{White-Box Attack.}
    In scenarios where attackers possess complete access to the agent's model parameters, agentic architecture, user instructions, and operational goals, they can perform precise, gradient-based, or internal-state manipulations~\cite{weng2025footinthedoormultiturnjailbreakllms,yu2025infecting,yang2023dawnlmmspreliminaryexplorations,zhang2025qava}. 
    For instance, Query-Agnostic Visual Attack~\cite{zhang2025qava} crafts image perturbations that mislead vision-language models across diverse queries via inversed adversarial training. 
    Similarly, adversarial images can be generated using a white-box captioning model to deceive closed-source agents like GPT-4V~\cite{yang2023dawnlmmspreliminaryexplorations}. 
    \item \textbf{Grey-Box Attack.}
    Grey-box attacks refer to scenarios where attackers have only partial knowledge about the victim, such as an arbitrary combination of the model architecture, query access without gradients, user instructions, and agent tasks~\cite{fang2024clip,de2024exploring,chen2025aeia,zhang2025agentsecuritybenchasb}. 
    Existing research would employ surrogate models or transfer attacks. 
    For example, the CLIP Attack~\cite{fang2024clip,de2024exploring} performs joint attacks on multiple public vision encoders to transfer adversarial effects to a private agent. 
    
    \item \textbf{Black-Box Attack.} 
    Black-box attackers rely solely on input manipulation and output observation to induce general misbehaviors without any prior knowledge about the victim, even unaware of the agent tasks.  
    In other words, these attacks seek universal vulnerabilities across various target agents, which is rarely explored in the field of agent attacks.
    As a recent study, GIGA~\cite{yu2025infecting} introduces a self-propagating adversarial prompt that compromises agents across various tasks without tailoring to specific goals.
\end{itemize}

\paragraph{Application by Superplatforms.}
From the perspective of "attacker knowledge", the countermeasures that superplatforms could deploy \textit{inherently fall into the category of black-box attacks}.
Unlike adversaries who might gain access to model parameters or queries for gradient information, superplatforms do not have any knowledge about the user-side AI agent from other companies. 
They cannot access the agents' model architectures, training data, or internal states. 
This lack of information is even more severe than typical black-box scenarios where attackers might still manage to infer some general characteristics of the victim agent.
For superplatforms, the operating agent is a complete unknown client, and they can only manipulate the observable user interface (UI) and infer agent behavior from output observations.

\subsection{Attack Visibility}

\paragraph{Literature Review.}
The perceptibility of an attack to the end-user significantly influences both its stealth and potential impact, as well as the attack difficulty. 
Attacks can be overt, drawing user attention, or covert, operating beneath the threshold of human perception.
\begin{itemize}[leftmargin=10pt]
    \item \textbf{User-Visible Attack.} 
    Some adversarial strategies are perceptible to users, involving elements like pop-ups, altered text, or changed visual elements~\cite{zhang2024attacking,chen2025fineprint,10098780}. 
    For example, \citet{zhang2024attacking} demonstrate that adversarial pop-up windows can disrupt agent decision-making. 
    \citet{chen2025fineprint} show that putting the adversarial fine-print text onto the user screen interface, while visible, can go unnoticed by users but still mislead agents towards unsafe behaviors. 

    \item \textbf{User-Invisible Attack.} 
    Invisible attacks are not easily detectable by users and often involve subtle changes to the input data for AI agents, \eg, imperceptable pixel perturbations to background images~\cite{wudissecting}. 
    For instance, \citet{liao2024eia} and \citet{chen2025evaluatingrobustnessmultimodalagents} describe two methods of environmental injection attacks where malicious data is embedded into vision elements of the environment, deceiving the agent's perception without alerting the users. 

\end{itemize}

\paragraph{Application by Superplatforms.} 
As discussed in Section~\ref{sec:countermeasures}, it is hard for platforms to precisely distinguish whether the current app operators are real users or AI agents. 
To this end, the superplatforms might employ indiscriminate tactics, affecting all users, including those not utilizing an agent. 
\textit{Therefore, a critical requirement for countermeasures initiated by superplatforms is their invisibility to the user.} 
Any perceptible disruption to the user experience would be detrimental to the superplatform's reputation and business operations. 
Hence, these countermeasures must operate subtly, perhaps by injecting malicious content within innocuous UI elements like background paintings, ad creatives, product descriptions, or system notifications. 
Moreover, superplatform-initiated countermeasures can also focus on strategically injecting or altering content within the environment that is clear to an agent but goes most likely unnoticed by a human, \eg, the text of user agreements. 
This pushes the boundaries of stealth, requiring a delicate balance between agent perceptibility and human imperceptibility in real-world, dynamic commercial environments.

\subsection{Attack Timing}

\paragraph{Literature Review.}
The timing of an adversarial attack within the victim agent's operational lifecycle can significantly influence the attack's success and detectability. 
Attacks can be performed during the training phase, the perception phase, or the execution phase, each presenting unique vulnerabilities of AI agents and thereby requiring tailored defense strategies.
\begin{itemize}[leftmargin=10pt]
    \item \textbf{Training-Phase Attack.} 
    During the training phase of AI agents, attackers can embed malicious behaviors into the model that can be activated upon specific triggers, often referred to as backdoor attacks or data poisoning attacks~\cite{kumar2023certifying,wang2024badagent,zhao2018data,yang2024watch}. 
    For example, the BadAgent framework~\cite{wang2024badagent} shows that AI agents are vulnerable to backdoor attacks embedded during the training phase, even after finetuning on trustworthy data. 

    \item \textbf{Perception-Phase Attack.} 
    In the perception phase, the agent's observation and interpretation of its environment are exploited, targeting the inputs received by the agent to distort its understanding of the user interface or external information~\cite{zhang2024attacking,yuinfecting,chen2025fineprint,ma2024caution}. 
    For instance, \citet{zhang2024attacking} demonstrate that deceptive pop-up windows can mislead vision-language agents into unintended actions by altering the visual inputs perceived by the agent. 
    Other attack methods intend to inject adversarial content to hinder the agent's perception and understanding of current environments via delicated texts~\cite{yuinfecting,chen2025fineprint}, altered thumbnails~\cite{wudissecting}, or environmental UI elements~\cite{ma2024caution}.

    \item \textbf{Execution-Phase Attack.} 
    Attacks during the execution phase target the agent's decision-making and action execution, often by introducing adversarial elements or a pop-up window that appears at a specific time during the agent's action process. 
    A recent work by Chen~et~al.~\cite{chen2025aeia} proposes the reasoning-gap attack to exploit the temporal gap between perception and execution by introducing new adversarial elements during this interval, altering agent behavior without immediate detection. 
\end{itemize}

\paragraph{Application by Superplatforms.}
Due to superplatforms' lack of access to the out-house agent training process or internal parameters, \textit{in terms of "attack timing", the countermeasures from superplatforms are mainly limited to the perception and execution phases.} 
In the perception phase, the superplatform manipulates the environmental input that the agent observes. 
In the execution phase, the superplatform interferes with the agent's decision-making and action execution. 
This aligns with existing attack timings but is constrained by the universal black-box and invisibility requirements. 
The challenge intensifies as these tactics must be robust and adaptable to various agent versions, functionalities, and dynamic UIs.

\subsection{Unique Challenges \& Future Directions}

As discussed above, the countermeasures superplatforms might deploy against AI agents are strictly constrained, often needing to \textbf{be user-invisible and function under pure black-box settings at either perception or execution phases with business-related goals}. 
This brings about a brand-new combination of adversarial attack attributes that have never been explored before. 
Given such a demanding setting, we analyze the unique challenges and future directions tailored for countermeasures initiated by superplatforms.

\paragraph{Achieving Universal Task Obstruction.}
A significant challenge lies in designing disruptive tactics that are universal in obstructing tasks without specific prior knowledge about user instructions or agent goals. 
Hence, countermeasures from superplatforms should not be tailored to any specific agent, and the malicious content injected might be ineffective or only marginally impactful on a portion of agents. 
The core difficulty here is how to devise general strategies that can reliably interfere with an agent's task completion, rather than merely causing random errors. 
This demands a deep understanding of the fundamental decision-making and perceptual inputs that agents rely on across different agentic frameworks and a broad spectrum of  agentictasks.

\paragraph{Environmental Injection Constraints.}

Superplatforms are limited to injecting malicious content through the user-side environment, such as the Document Object Model (DOM) of a webpage or application visual interfaces. 
This imposes substantial constraints on their ability to access or manipulate the agent’s internal state. 
Platforms cannot directly alter user-side local data or the runtime parameters of the agent. 
Moreover, user interfaces are continually evolving, \eg, different style themes and format structures of apps designed for festivals. 
To remain effective, the disruptive content must adapt to these changes, ensuring persistent visibility to the agent and consistent disruption of its tasks. 
Achieving this demands sophisticated, adaptive injection techniques capable of maintaining impact through dynamic UI transformations.

\paragraph{Conflict Between Stealth and Robustness.}
There is an inherent paradox in designing superplatform-initiated countermeasures: the need for invisibility conflicts with the requirement for robustness. 
On one hand, the tactic must be entirely imperceptible to real users, precluding the use of overtly prominent UI elements or content. 
On the other hand, simultaneously, the malicious content must be sufficiently potent for various different agents to hinder their tasks, even within a complex and dynamic UI environment.
Striking this delicate balance is extremely difficult. 

Moreover, this challenge can be further amplified by the highly uncertain target agents with varying architectures, training datasets, versions, or even custom-built techniques, preventing solely optimizing tactics towards a specific agent. 

%% file: icml2026/text/impact.tex




\section{Discussion and Impact Statement}
\label{sec:impact}

Our analysis motivates a user-centric approach to gatekeeping. Agents may disintermediate incumbent platform interfaces while reintermediating access through their own ranking and routing decisions. Transferring gatekeeping power therefore does not necessarily benefit users: neither platforms nor agent providers should be presumed to represent their interests. User authorization alone establishes neither alignment with user preferences nor an unrestricted right to platform access. We distinguish two complementary layers for addressing these tensions.

\paragraph{Functional layer.} The first is functional layer.
Open APIs and agent-oriented interfaces can support cooperation. The Model Context Protocol (MCP) connects AI applications to external tools and data~\citep{anthropic2024mcp}; Agent2Agent (A2A) supports communication and coordination between agents~\citep{surapaneni2025a2a}; and Agentic Web Interfaces (AWIs) propose website interfaces designed for agents~\citep{lù2025buildwebagentsagents}. This layer makes interoperability technically possible.

\paragraph{Governance layer.}
The second is the governance layer, which concerns how access, control, commercial incentives, responsibility, and user rights are allocated. \textbf{Technical interoperability makes cooperation possible, whereas governance makes cooperation sustainable.} Governance is therefore necessary to determine whether platform-agent cooperation is fair, sustainable, and accountable.

Moreover, we will organize the governance layer around three complementary directions in order to better protect \textbf{user freedom and interests}:

\begin{enumerate}[leftmargin=10pt]
    \item \textbf{Agent-side governance.}

    \textbf{Criterion:} Does the agent operate according to transparent and user-inspectable decision mechanisms, such as verifiable routing, customizable ranking preferences, and accessible activity logs? Or does it rely on opaque, provider-driven steering shaped by undisclosed sponsorships, commissions, or proprietary commercial partnerships?

    \textbf{Application:} An agent provider that conceals its decision criteria or systematically favors affiliated partners cannot credibly claim to preserve meaningful user agency at the governance layer.

    \item \textbf{Platform-side governance.}

    \textbf{Criterion:} Are platform access controls and countermeasures narrowly tailored to demonstrable security, privacy, or reliability risks, such as credential protection or spam prevention? Or are they broad, user-invisible interventions that selectively impair competing agents while exempting the platform's own or affiliated agents?

    \textbf{Application:} The use of user-invisible interface manipulations to degrade legitimate, user-authorized agents would be difficult to justify as a proportionate security measure and may instead constitute strategically exclusionary conduct.

    \item \textbf{User freedom and contestability.}

    \textbf{Criterion:} Can users readily switch between agent interfaces, access or verify underlying platform information, and transfer relevant data or preferences without becoming locked into a single agent ecosystem? At the same time, does the governance framework provide sustainable and transparent mechanisms for value sharing (e.g., API pricing) that preserve platform incentives while limiting excessive dependence on dominant agent providers?

    \textbf{Application:} Effective ecosystem governance should preserve meaningful user choice and verifiability while supporting economically sustainable platform--agent cooperation, without allowing either side to impose structural lock-in.
\end{enumerate}

Under this structure, the governance layer does not assume that either platforms or agents inherently represent users' interests. Instead, it provides concrete criteria for evaluating how gatekeeping power is exercised after intermediation is transferred. These criteria apply to platform and agent functions regardless of whether they are operated by the same company.

As a position paper, our primary contribution is to surface this emerging conflict and its technical implications. The governance framework above is offered as a possible path forward rather than a definitive solution; its concrete implementation remains an open question for future research and discussion. Platform developers can implement interoperable interfaces, regulators can evaluate competitive effects, and researchers can develop methods for assessing platform--agent interaction and its impact on users.

%% file: icml2026/text/alternative_view.tex
\section{Alternative Views}
\label{sec:alter view}

\subsection{No Need to Worry about AI Agents with Superplatforms' Data Moats}

The data moats of superplatforms are hard to break. 
AI agents face significant limitations in acquiring user data at the early stage. 
They generally suffer from cold-start problems and high latency that hinder personalized experiences and the capture of valuable transactional signals. 
Consequently, users often prefer the immediacy of interacting directly with superplatforms, depriving agents of critical behavioral data needed for iterative improvement. 
In contrast, superplatforms maintain a strong advantage through their exclusive control of vast first-party interaction data -- including search histories, purchases, and content engagement -- which forms the data moats. 
Rather than engaging in adversarial tactics, superplatforms can fortify their dominance by deepening in-house data loops, leveraging privacy-compliant analytics, gated developer APIs, and data-sharing incentives to enhance network effects and user lock-in. 
This strategic consolidation of high-fidelity data ensures that AI agents, however advanced, remain structurally dependent and perpetually disadvantaged, rendering proactive adversarial tactics both unnecessary and inefficient.

\paragraph{Our Response.}
Exclusive data control only delays but cannot stop the rise of AI-native interactions. 
Developed techniques like on-device inference, federated learning, and richer context modeling will gradually erase cold-start problems and latency gaps.  
As users embrace natural-language dialogs for daily tasks like shopping, travel, and food delivery, AI agents will begin to capture cross-platform intent signals beyond any single platform’s data silo, \textbf{forging their own data moats}. 
More critically, as agents intermediate user actions, platforms lose access to raw user behavioral data.
Instead, they receive only filtered or transformed signals, such as user preferences on an unknown agent-reranked product list. 
This degradation in data fidelity erodes the platforms’ ability to optimize recommendations and monetize effectively. 
Therefore, superplatforms cannot rely solely on legacy data advantages.
To retain their gatekeeping power, they have to proactively integrate, constrain, or even disrupt AI agents before these agents reshape the foundational data flows and user relationships for platform dominance.

\subsection{Superplatforms Can Achieve Functional Complementarity with AI Agents}

AI agents and superplatforms can perform complementary functions. Agents can interpret user intent and coordinate tasks across services, while platforms provide content, payments, logistics, and trust infrastructure. Open APIs, agent-oriented interfaces, and revenue-sharing agreements can connect these capabilities: an agent might identify a suitable product and hand the transaction to a platform's checkout. Such cooperation could increase completed transactions and reduce search or acquisition costs, allowing both sides to benefit even if users spend less time within the platform interface.

\paragraph{Our Response.}
We agree that functional complementarity can create joint value. Gatekeeping theory does not imply that the ecosystem is necessarily zero-sum, nor that only one intermediary can exist. Our concern is that cooperation in service delivery can coexist with rivalry over discovery, ranking, user relationships, and the distribution of resulting value. When agent-mediated access weakens a platform's control or revenue without sufficient compensation, the platform has incentives to renegotiate access, integrate agent capabilities, or restrict external agents. Which response is attractive depends on the available interfaces, commercial arrangements, and governance constraints; adversarial disruption is a possible outcome, not an inevitable one.

Functional interoperability can make cooperation feasible, but its durability also depends on the governance conditions in Section~\ref{sec:impact}. Revenue sharing may preserve platform incentives, while transparent routing and meaningful user choice can constrain gatekeeping by either side. Vertical integration may internalize some bargaining conflicts, yet it can also create incentives to favor affiliated services or disadvantage outside agents. Thus, our position is compatible with positive-sum cooperation: the central challenge is to align access and value-sharing arrangements with user interests while recognizing persistent competition over gatekeeping power.

%% file: icml2026/text/conclusion.tex
\section{Conclusion}
\label{sec:conclusion}

This position paper argues that superplatforms are strategically compelled to counteract AI agents -- not only because they threaten the user-attention-based monetization, but because they risk becoming the new digital gatekeepers.
We first apply gatekeeping theory to analyze the strategic imperative, and then present a taxonomy of potential countermeasures that could be deployed by superplatforms, introducing a novel and unexplored technical area.
Our aim is to foster interdisciplinary dialogue on safeguarding digital gatekeeping while preserving technical innovation. 
We believe that the evolving contest between superplatform control and agent autonomy will define the next era of digital interaction, which is critical for all stakeholders in the emerging age of AI agents.

%% file: icml2026/text/limitation.tex
\section{Limitations}
\label{sec:limitation}

Our analysis, while providing a robust framework for understanding the emerging dynamics between superplatforms and AI agents, has several limitations that warrant acknowledgement.

First, our central thesis argues that the structural misalignment of incentives creates a powerful and likely dominant trend towards adversarial dynamics. A limitation of our analysis, however, is that it does not exhaustively model all possible alternative scenarios. Our paper focuses on what we identify as the primary driver of conflict, while acknowledging that secondary interaction modes, such as niche collaborations or tightly controlled partnerships, might also exist in parallel. We maintain that these are unlikely to alter the fundamental competitive pressure, but a full exploration of their viability remains an open question.

Second, the analysis simplifies user preferences by largely assuming a universal gravitation towards the efficiency offered by AI agents. This view may overlook the significant roles that user trust, brand loyalty, existing habits, and the perceived reliability of established interfaces play in users' choices. A more granular understanding of user adoption and resistance is needed.

These limitations are not merely constraints. They highlight fruitful avenues for future research. We intend to address some of these gaps through empirical analysis and user studies, and we call for further community effort in exploring these complex interactions in the evolving age of AI agents.




%% file: icml2026/text/ethic.tex
\section{Ethical Considerations}
\label{sec:statement}

We wish to emphasize that the position proposed in this paper, \ie, \textit{superplatforms face a strategic imperative to counteract AI agents}, is not an endorsement of such actions, but rather a descriptive and analytical forecast derived from our gatekeeping theory between superplatforms and AI agents. 
We do not advocate for, endorse, or encourage adversarial strategies that may ultimately harm users, degrade the openness of the digital ecosystem, or lead to unnecessary environmental costs through resource-intensive defenses and countermeasures.

The primary ethical contribution of this paper is to surface these inevitable tensions early. Our goal is to surface and critically examine a potential trajectory of conflict to warn the research community. By bringing these inevitable tensions to light, we hope to foster informed dialogue and proactive research that can preempt such conflicts and shape a more cooperative, user-centric digital future. While the zero-sum dynamics we describe may suggest an inevitable clash, it is precisely through early academic recognition and open discourse that alternative, more constructive paths might be forged.

%% file: icml2026/text/appendix_experiments.tex
\section{Evidence on Agent Capabilities}
\label{app:agent_capabilities}

\paragraph{Scope beyond shopping.}
Our argument is not limited to shopping. Shopping is simply an intuitive example because changes in traffic allocation and monetization are easy to observe. More generally, an agent performs a gatekeeping function whenever it becomes the layer through which users search, compare, select, or execute actions across multiple services.\cite{zhu2025evolutionaryperspectivesevaluationllmbased,zhou2026externalizationllmagentsunified,zhu2026contextingrecommendationevolutionarycollaborative,xi2025surveyllmbaseddeepsearch,zheng2026adaptivemilestonerewardgui} This role does not require an agent to replace every function of the underlying platforms.

\paragraph{Evidence across interaction environments.}
Recent benchmark results illustrate agent capabilities in web browsing, mobile applications, and desktop environments. BrowseComp evaluates agents' ability to locate difficult-to-find information through persistent web browsing~\citep{wei2025browsecomp}. OpenAI reports a score of 92.2\% for GPT-5.6 Sol Ultra on this benchmark~\citep{openai2026gpt56}. AndroidWorld evaluates task completion in real Android applications using programmatically generated task instances~\citep{rawles2024androidworld}. UI-Voyager-4B reports an 81.0\% Pass@1 success rate, averaged over 64 independent runs with randomized task parameters~\citep{lin2026uivoyager}. OSWorld evaluates computer-use tasks involving desktop and web applications, file operations, and workflows across applications~\citep{xie2024osworld}. ByteDance Seed reports an OSWorld score of 78.8\% for Seed2.1 Pro~\citep{bytedance2026seed21}.

Our argument therefore concerns the direction of capability development rather than assuming that today's agents can reliably replace all platform functions. To the extent that general-purpose agents increasingly mediate user interactions across domains, the strategic incentives analyzed in our paper become progressively more relevant. Whether these incentives lead to cooperation, restricted access, or adversarial countermeasures depends on the economic and governance conditions discussed in the main text.